\documentclass{article}

\usepackage{PRIMEarxiv}

\usepackage[utf8]{inputenc} 
\usepackage[T1]{fontenc}    
\usepackage{hyperref}       
\usepackage{url}            
\usepackage{booktabs}       
\usepackage{amsfonts}       
\usepackage{nicefrac}       
\usepackage{microtype}      
\usepackage{xcolor}         
\usepackage{natbib}
\usepackage{graphicx} 
\usepackage{amsmath} 

\usepackage[utf8]{inputenc} 
\usepackage[T1]{fontenc}    
\usepackage{hyperref}       
\usepackage{url}            
\usepackage{booktabs}       
\usepackage{amsfonts}       
\usepackage{nicefrac}       
\usepackage{microtype}      
\usepackage{lipsum}
\usepackage{fancyhdr}       
\usepackage{graphicx}       
\graphicspath{{media/}}     
\DeclareUnicodeCharacter{015E}{\c{S}} 
\DeclareUnicodeCharacter{0131}{\i}     
\DeclareUnicodeCharacter{0307}{\textperiodcentered} 

\title{How does a Language-Specific Tokenizer affect LLMs?}

\author{%
  Jean Seo, Jaeyoon Kim, SungJoo Byun, Hyopil Shin\\
  Seoul National University\\
  \texttt{\{seemdog, toscour345, byunsj, hpshin\}@snu.ac.kr} \\
}

\begin{document}
\maketitle

\begin{abstract}
The necessity of language-specific tokenizers intuitively appears crucial for effective natural language processing, yet empirical analyses on their significance and underlying reasons are lacking. This study explores how language-specific tokenizers influence the behavior of Large Language Models predominantly trained with English text data, through the case study of Korean. The research unfolds in two main stages: (1) the development of a Korean-specific extended tokenizer and (2) experiments to compare models with the basic tokenizer and the extended tokenizer through various Next Token Prediction tasks. Our in-depth analysis reveals that the extended tokenizer decreases confidence in incorrect predictions during generation and reduces cross-entropy in complex tasks, indicating a tendency to produce less nonsensical outputs. Consequently, the extended tokenizer provides stability during generation, potentially leading to higher performance in downstream tasks.

\end{abstract}

\section{Introduction}

The training data for Large Language Models (LLMs) today is heavily skewed towards English and code. For instance, Llama-2 \citep{touvron2023llama} from Meta-AI\footnote{https://www.meta.ai/} is trained with 89.7\% English, 8.38\% code, while other languages make up the remaining 2\%. Although these models are claimed to be multilingual, the strong emphasis on English affects the tokenizer's ability to effectively process languages other than English, consequently impacting the model's multilingual performance. As opposed to the tokenization of English text where text is segmented into appropriate units that capture linguistic information well, they often segment other languages—especially non-alphabetic ones—into bytes. Excessive byte-level segmentation can lead to issues such as limiting the maximum input and output length. Additionally, a byte token can be decoded into subwords of different languages depending on the neighboring tokens. This makes efficient representation embedding challenging, as a single entry may need to absorb too much information.

Moreover, the optimal tokenizing strategy for each language can vary depending on linguistic features, such as whether the language is agglutinative or inflective. This means the tokenizer of a certain LLM might not be fundamentally suitable for some languages. However, using a completely new tokenizer for a specific language is challenging because it would require retraining the language model from scratch. Therefore, many language-specific fine-tuned LLMs use tokenizer extension, incorporating a larger vocabulary size and additional merge rules for the target language. While extending the tokenizer also necessitates further training to align the embeddings, it is much more efficient than training the model from scratch with an entirely new vocabulary.

The necessity of language-specific tokenizers is supported by works such as \citet{unpacking, arabic}, which demonstrate that tokenization strategy and vocabulary size affect the performance of downstream tasks. However, deeper analysis is needed to understand exactly how tokenizers impact LLM performance. Such insights could explain why different tokenizers lead to varying model performance and increase the explainability of black-box LLMs. This, in turn, would provide evidence to inform decisions when building better models in the future.

In this research, we conduct an in-depth analysis of the effect language-specific tokenizers have on LLM performance. For an intrinsic analysis, we employ a Next Token Prediction (NTP) task with varying levels and develop corresponding metrics to evaluate the influence of different tokenizers. We experiment with Korean, a non-English, non-alphabetic language. The experimental results show that the extended tokenizer produces more sensible outputs more stably, with lower confidence levels in incorrect token generation and lower cross entropy in complex tasks. This underscores the importance of tokenizer extension, supporting claims from previous works that language specific tokenizers lead to elevated downstream task performance \citep{arabic, llm_tokenizer}.

The contributions of this study are:

\begin{itemize}
    \item Intrinsically analyzing the effect of language-specific tokenizers on how LLMs work
    \item Proposing a novel framework for the intrinsic evaluation of tokenizers with Next Token Prediction task, using accuracy, confidence level, and  cross entropy loss as metrics
\end{itemize}

\section{Background and Related Work}

Traditionally, Language Model (LM) tokenizers were primarily optimized for English, often resorting to byte-level segmentation for non-English languages. Segmenting text into byte-levels has a critical drawback: it greatly limits the maximum input and output length of the model. To address this issue, recent LLMs such as Llama-3 \citep{llama3modelcard} and Gemma \citep{gemmateam2024gemma} adopt a different tokenization strategy, exponentially expanding the vocabulary size to 128,256 and 256,128 tokens, respectively. Therefore, a common approach has been developed to extend an LLM tokenizer to be specialized in one additional language besides English, expanding vocabulary and merge rules of only this target language. This approach is demonstrated by several Korean-extended models \citep{llama3openko, gemma_ko_7b, l._junbum_2023, kim2024efficient}.

Prior research on tokenizers has predominantly focused on evaluating their effectiveness for specific languages \citep{How_good_is, turkish} or domains \citep{domain, domain2}. Tokenization evaluation is essential for enhancing common knowledge, analyzing linguistic analysis, distinguishing processing types, and improving tokenizer design based on the presented dimensions \citep {1998TowardsTE}. Nonetheless, there is a clear lack of research directly examining their influence on LLMs. The few studies about evaluating the effect of tokenizers on LLMs include \citet{unpacking, japanese}, which are based on the downstream performance. \citet{arabic} examined how different tokenization strategies and vocabulary sizes affect the performance of Arabic language models in downstream tasks. \citet{llm_tokenizer} studied the influence of tokenizer choice on LLM downstream performance by training mono- and multilingual LLMs. However, it is uncertain whether downstream task performance can accurately measure the effectiveness of tokenizers. There remains a need for intrinsic and analytical studies to understand why performance differences occur. Such investigations are imperative to mitigate the opacity of black-box models and provide a rationale for decision-making towards the development of superior models.

\begin{figure}[h] 
    \centering
    \includegraphics[width=0.9\columnwidth]{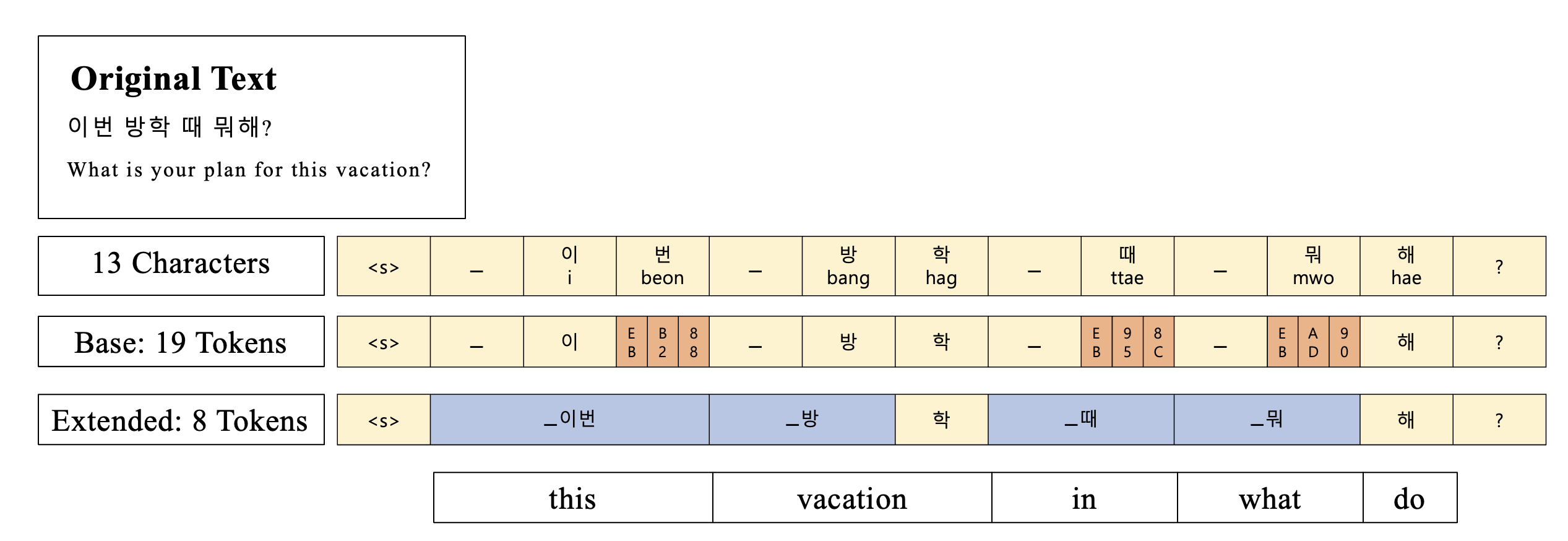}
    \caption{An example comparing how the Llama-2 base tokenizer and our extended tokenizer process the same sentence. The example sentence consists of 13 characters, including the special token <s> and whitespace. The base tokenizer segments 3 of these characters into bytes, resulting in a total of 19 tokens. In contrast, our extended tokenizer appropriately tokenizes the sentence into meaningful units of subwords, resulting in only 8 tokens.}
    \label{fig:tokenized}
\end{figure}

\section{Extended Tokenizer}

Well-known open-source models, such as  Mistral \citep{jiang2023mistral} Llama-2 \citep{touvron2023llama}, Yi \citep{ai2024yi}, MT5 \citep{xue2021mt5}, Giraffe \citep{niemeyer2021giraffe}, Solar \citep{kim2024solar} use SentencePiece BPE. Meanwhile, models including Falcon \citep{almazrouei2023falcon}, Llama-3- \citep{llama3modelcard}, GPT-Neox \citep{black-etal-2022-gpt}, Bloom \citep{workshop2023bloom}, Qwen \citep{bai2023qwen} use ByteLevel BPE as their tokenization strategy. This shows that SentencePiece BPE and ByteLevel BPE are currently the most common tokenization methods for LLMs.

In this study, we employ TinyLlama \citep{zhang2024tinyllama} to investigate the effect of language-specific tokenizers. TinyLlama, which has 1.1 billion parameters, is relatively small and trained with barely any amount of Korean text. It uses the Llama-2 tokenizer which is based on SentencePiece BPE, combining SentencePiece \citep{kudo2018sentencepiece} with the BPE algorithm. The main advantage of SentencePiece BPE is that predicted tokens can be directly joined into sentences. Additionally, it offers the Byte Fallback option, which processes unknown input sequences into bytes based on the UTF-8 encoding of characters, enhancing its ability to handle unknown tokens.

Despite its ability to handle unseen tokens by splitting them into bytes, SentencePiece BPE can cause issues when processing languages underrepresented in the training data or vocabulary. This is because a byte can be decoded into various words or subwords depending on its context, making efficient learning of representations difficult. Byte-level segmentation also reduces the model's maximum input and output sequence length.

To address these issues, we build a language-specific tokenizer through vocabulary and merge rule extension. For our case study on Korean, we train a SentencePiece BPE tokenizer on Korean texts and append the new Korean vocabulary and merge rules to those of TinyLlama's tokenizer. Figure \ref{fig:tokenized} shows that the extended tokenizer segments Korean sentences into bytes less frequently and into larger subwords with lexical and grammatical meaning. Furthermore, Table \ref{tab:unk} demonstrates the superior performance of the extended tokenizer compared to other LLM tokenizers.

\begin{table}[t]
\centering
\resizebox{0.9\columnwidth}{!}{%
\renewcommand{\arraystretch}{1.3}
\begin{tabular}{c|c|c|c}
\toprule[1.3pt]
\textbf{Tokenizer}                & \textbf{Extension} & \textbf{\% of Unknown Tokens} & \textbf{Average Number of Tokens} \\ \toprule[1.3pt]
\textbf{Mistral-7B-Instruct-v0.1} & \textbf{N}         & 14.36                       & 242.71                         \\ 
\textbf{Mistral-ko-7B-v0.1}       & \textbf{Y}         & 1.0                         & 128.56                         \\ \hline
\textbf{Llama-2-7b}               & \textbf{N}         & 48.61                      & 325.27                         \\
\textbf{Llama-2-ko-7b}            & \textbf{Y}         & 0.98                       & 128.92                   
                              \\
\textbf{Llama2-extended(Ours)} & \textbf{Y}         & \textbf{0.96}                            & \textbf{125.91}                               \\ \toprule[1.3pt]
\end{tabular}%
}
\caption{
The percentage of unknown tokens and the average number of tokens per sentence in our Korean test sentences evaluated across various tokenizers. Unknown tokens refer to tokens that are segmented into bytes. We used different text data for tokenizer evaluation than the text data used for tokenizer extension. Extended tokenizers show a significantly lower percentage of unknown tokens, demonstrating less text segmented into bytes, and a reduced average number of tokens, indicating a longer maximum input and output length. Amongst the extended tokenizers, our Llama2-extended tokenzier shows the least percentage of unknown tokens and the shortest average number of tokens of sentences.}
\label{tab:unk}
\end{table}

\section{Next Token Prediction}

In order to deeply analyze the effect of tokenizers on LLMs, we introduce a NTP(Next Token Prediction) task along with corresponding evaluation metrics. This approach transcends mere observation of downstream task performances, enabling a more comprehensive analysis of tokenizer efficacy, possibly explaining how LLMs with certain tokenizers yield better performance compared to others.

\subsection{Task}

Unlike Encoder-only models such as BERT \citep{devlin2019bert} which refer to both preceding and subsequent tokens to predict masked tokens, LLMs rely solely on previous tokens for causal language modeling. Therefore, for the NTP task, we meticulously curated test sentences featuring a target token, ensuring that only one answer is possible given the preceding tokens. Three human annotators manually went through all test sentences and sorted out the final ones with masked tokens that admit only one possible answer. Our Next Token Prediction task comprises six distinct tasks, categorized into two difficulty levels (easy and hard) and three target units (token, character, and word).

\subsubsection{Difficulty}

\begin{figure}[h] 
    \centering
    \includegraphics[width=0.9\columnwidth]{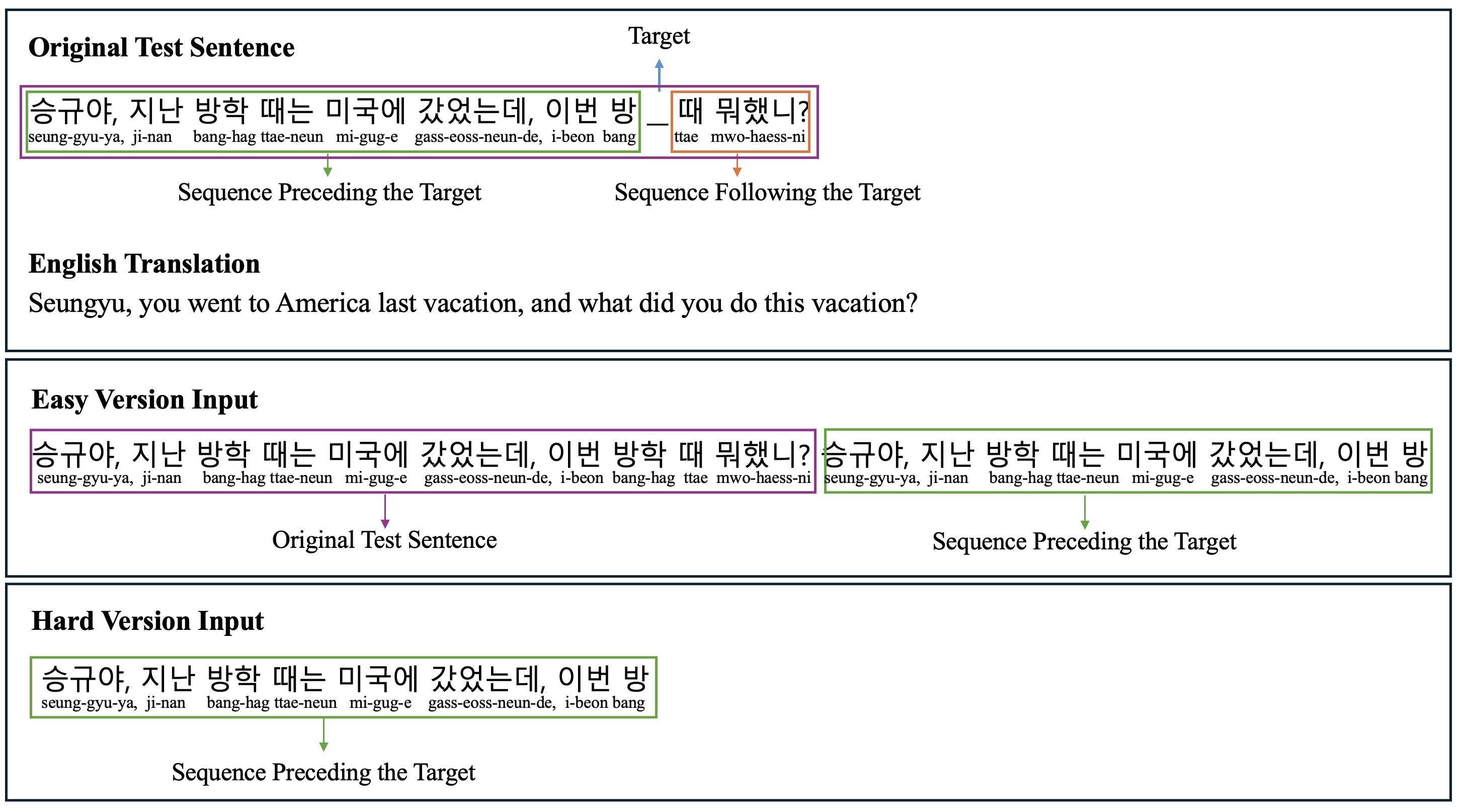} 
    \caption{Example of the NTP task input in easy and hard version respectively. In the easy version, the answer is already provided to the model since the input includes the original test sentence. In contrast, in the hard version, only the sequence preceding the target is provided, making it more difficult for the model to predict the correct token.}
    \label{fig:difficulty}
\end{figure}

The difficulty level of the NTP task depends upon the amount of input text provided to the model. Each test sentence is divided into three segments: the sequence preceding the target, the target itself, and the sequence following the target. In the easy version, the full sentence and the sequence preceding the target are provided as input, whereas only the sequence preceding the target is given in the hard version. Given that the model has already been exposed to the answer in the easy version, we presume predicting the target token to be considerably simpler than in the hard version. Further details regarding the two versions of the NTP task can be found in Figure ~\ref{fig:difficulty}.

\subsubsection{Target Unit}

\begin{figure}[h] 
    \centering
    \includegraphics[width=0.85\columnwidth]{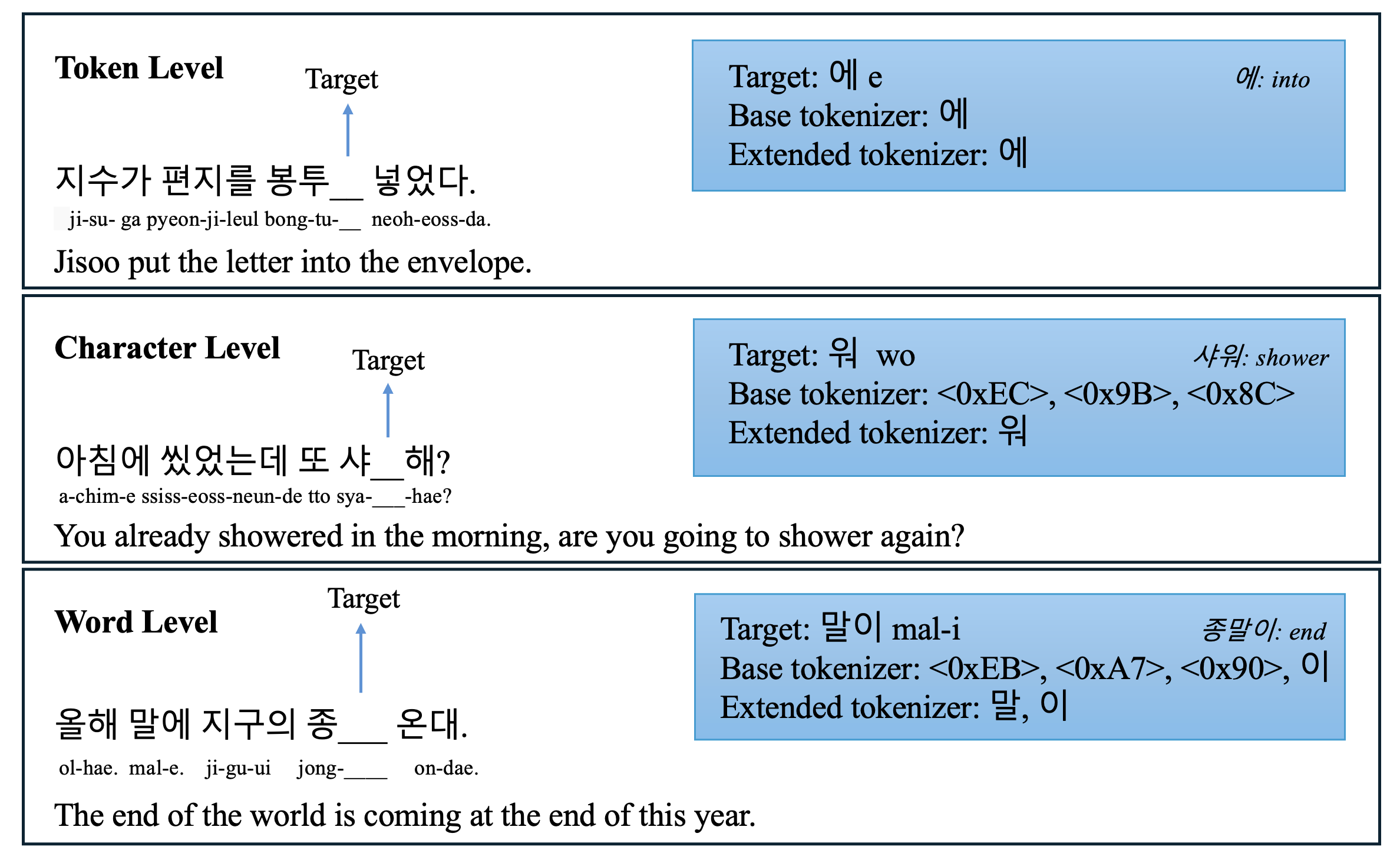} 
    \caption{Example of the NTP task input in the three different target units. At the token level, the target is tokenized into a single token by both tokenizers. At the character level, the target is segmented into 3 tokens by the base tokenizer but is treated as a single token by the extended tokenizer. At the word level, both tokenizers split the target into multiple tokens.}
    \label{fig:unit}
\end{figure}

In our comparison between two distinct tokenizers, the base tokenizer and the extended tokenizer, we categorize the target unit into three types: (1) token level, where both tokenizers tokenize the target into a single token; (2) character level, where the extended tokenizer tokenizes the target into a single token while the base tokenizer splits it into multiple tokens; and (3) word level, where both tokenizers tokenize the target into several tokens. Figure ~\ref{fig:unit} provides a detailed illustration of these three variations in the target units for the NTP task.

\subsection{Evaluation Metric}

To assess the intrinsic impact of tokenizers in LLMs through the NTP task, we employ three metrics: accuracy, confidence level, and cross-entropy loss.

\subsubsection{Accuracy}

We assess whether the model correctly predicts the target in a binary manner. When the target consists of a single token, determining accuracy is straightforward, as it can be classified as either correct or incorrect with ease. However, if the target comprises multiple tokens, we consider it correct only if the model predicts all tokens accurately. For instance, if the target is tokenized into three tokens and the model correctly predicts two tokens but incorrectly predicts one, the model's prediction is deemed incorrect. Evaluating the model's predictions binaryly, we compute the average accuracy across all test sentences.

\subsubsection{Confidence Level}

Building upon prior research that incorporates the notion of confidence level in deep learning, such as \citet{inoue2019adaptive}, we examine the model's confidence level during the generation of each target token. This approach allows for a more intrinsic evaluation and analysis of the effect of tokenizers on LLMs. Similar to \citet{schuster2022confident, niehues2019modeling}, where confidence level is used specifically in Language Modeling, we consider the probability calculated using the softmax layer as the confidence level at each time step. Formally, given an input token sequence \( X = x_1, \ldots, x_{i-1} \), the model predicts the next token \( x_i \) by computing the probabilities of all possible tokens in the vocabulary \( V \). These probabilities are obtained using the softmax function applied to the logits (the raw scores) produced by the model. Let \( z_j \) denote the logit corresponding to the \( j \)-th token in the vocabulary \( V \), where \( j \in \{1, \ldots, |V|\} \). The softmax probability for the \( j \)-th token is given by:

\[
P(x_i = j \mid X) = \frac{\exp(z_j)}{\sum_{k=1}^{|V|} \exp(z_k)}
\]

The confidence level at each time step, which is the maximum softmax probability, is then calculated as:

\begin{equation}
\text{Confidence}(X) = \max_{j \in \{1, \ldots, |V|\}} P(x_i = j \mid X) = \max_{j \in \{1, \ldots, |V|\}} \frac{\exp(z_j)}{\sum_{k=1}^{|V|} \exp(z_k)}
\end{equation}



To ensure more fair and accurate comparison between tokenizers with varying vocabulary sizes, we conduct normalization and additionally compare the normalized confidence levels. This necessity stems from the inherent characteristics of the softmax layer, wherein larger embedding sizes (or vocabulary sizes in this context) often lead to decreased probabilities assigned to each predicted token. To normalize the confidence level at time step \(t\), we first calculate the average confidence level of the previous \(t-1\) time steps:

\[
\overline{\text{Confidence}}(t-1) = \frac{1}{t-1} \sum_{k=1}^{t-1} \text{Confidence}(k)
\]

Then, we divide the confidence level at the current time step by this average:

\[
\text{NormalizedConfidence}(t) = \frac{\text{Confidence}(t)}{\overline{\text{Confidence}}(t-1)}
\]

By using the normalized confidence level, we can compare the effects of the base and extendend tokenizers while minimizing the impact of differing vocabulary sizes.

\subsubsection{Cross Entropy Loss}

Another metric we employ to compare the impact of different tokenizers on LLMs is Cross Entropy Loss. Cross Entropy Loss can be intuitively interpreted as a measure of how uncertain the model is when predicting the next token.

\section{Experiments}

\subsection{Models and Data}
\label{sec:exp}

To evaluate the isolated impact of the extended tokenizer, we train two models with all settings identical except for the tokenizer. Due to cost issues, we utilize a relatively small LLM, Tinyllama which has 1.1 billion trainable parameters as the base model. Our primary objective is to assess the influence of a tokenizer featuring a language-specific extended vocabulary—in this case, Korean—thus we compile a bilingual Korean-English pretraining dataset. Additionally, as our focus is on observing intrinsic effects rather than downstream task performance, we opt for a pretrained model instead of instruction-tuned model. The dataset comprises Korean news data from the Modu Corpus\footnote{https://kli.korean.go.kr/corpus/main/requestMain.do?lang=en}, English data from Wikipedia\footnote{https://www.wikipedia.org/}, and Alpaca dataset\citet{alpaca}. Given that the dataset used to train Tinyllama\citep{cerebras2023slimpajama, li2023starcoder} barely includes any Korean and our aim is to scrutinize the intrinsic impact of tokenizers on the LLM's ability to generate Korean, we maintain a dataset ratio of Korean to English of 9:1, enabling the model to become more adept with Korean. Furthermore, to investigate whether the quantity of training data correlates with the intrinsic ability of Korean token generation, we save intermediate checkpoints and evaluate them accordingly.

\subsection{Experimental Settings}

Each model was trained using identical hyperparameters: batch size = 4, linear learning rate scheduler, and weight decay = 0.01. Training was conducted using 3 A100 GPUs, with each checkpoint taking approximately 27 hours to be trained.

\section{Result and Analysis}

\begin{figure}[h] 
    \centering
    \includegraphics[width=0.7\columnwidth]{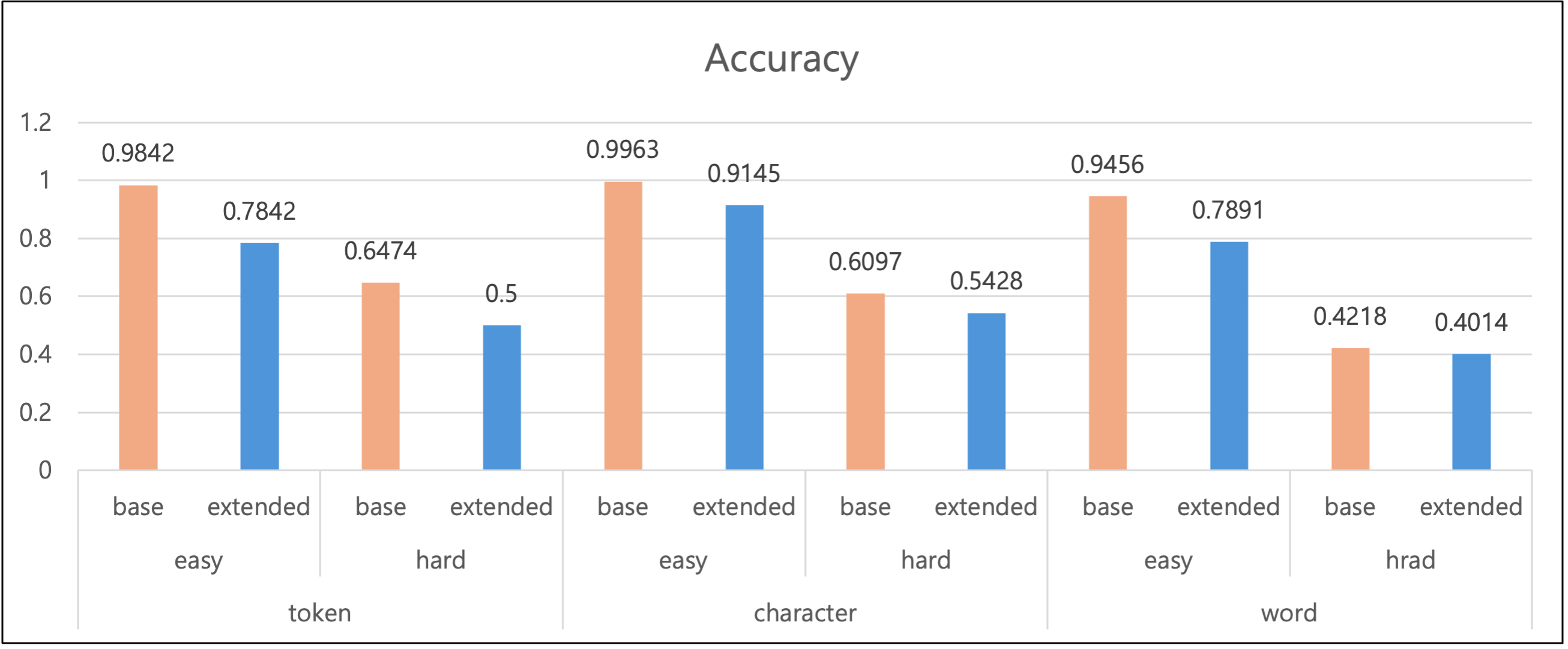} 
    \caption{Accuracy}
    \label{fig:accuracy}
\end{figure}

\begin{figure}[h] 
    \centering
    \includegraphics[width=0.7\columnwidth]{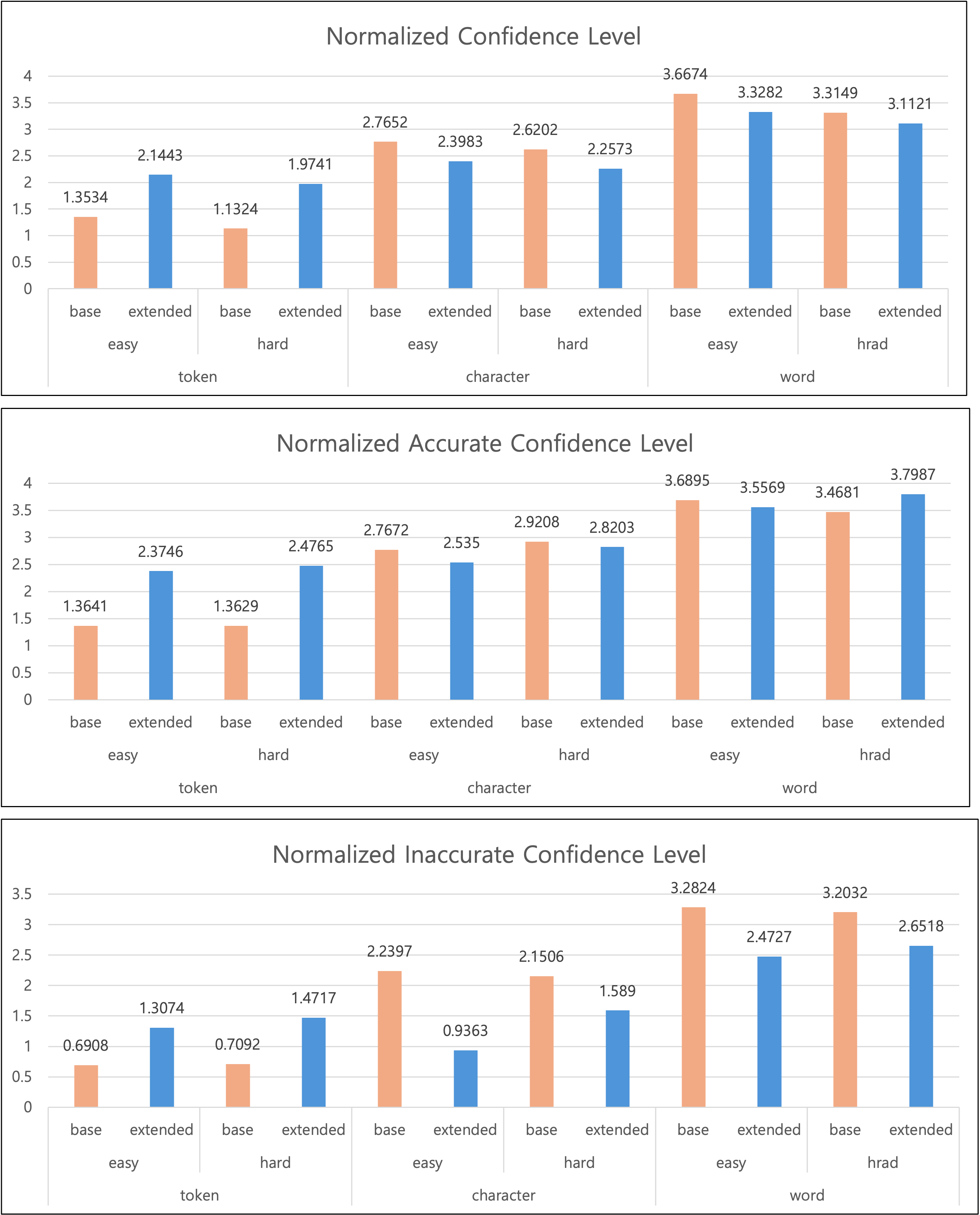} 
    \caption{Normalized Confidence Level}
    \label{fig:cl}
\end{figure}

\begin{figure}[h] 
    \centering
    \includegraphics[width=0.7\columnwidth]{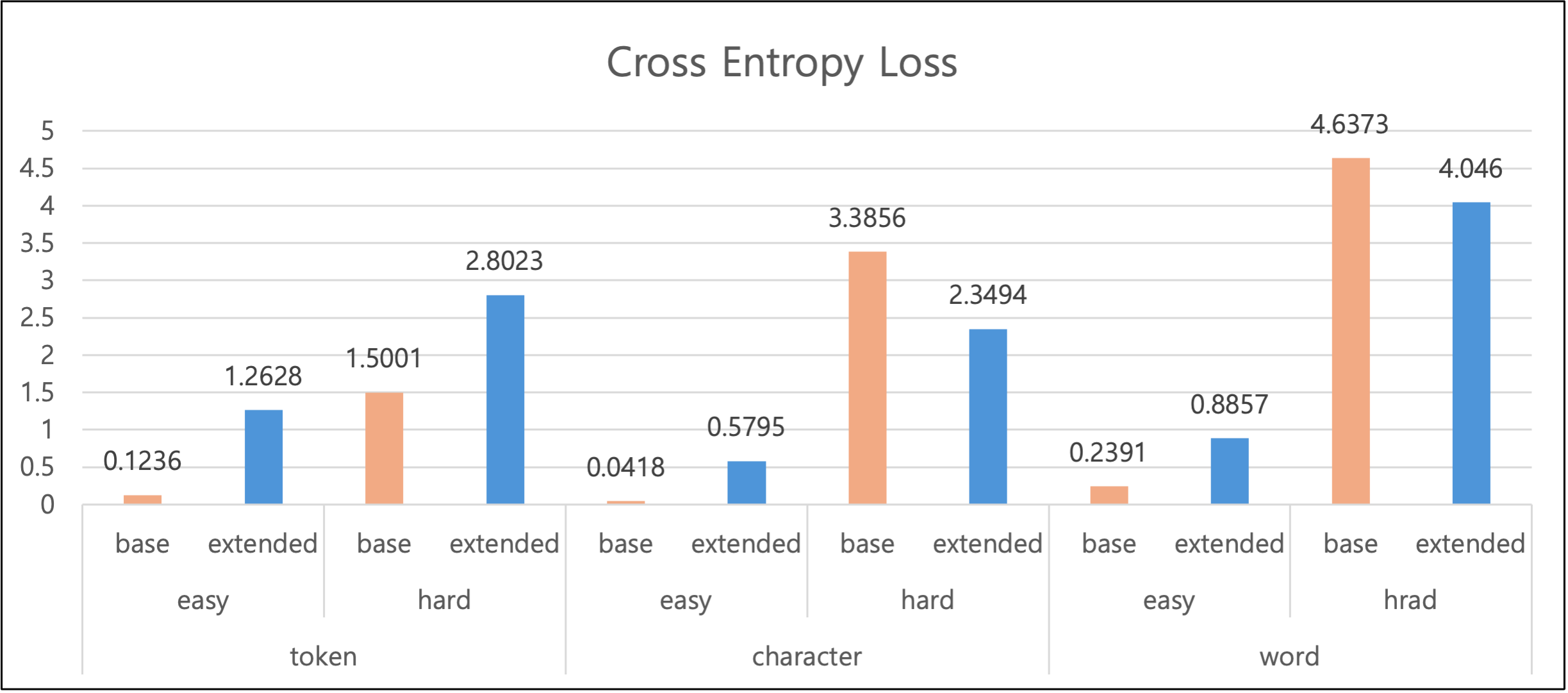} 
    \caption{Cross Entropy Loss}
    \label{fig:celoss}
\end{figure}

As mentioned in Section \ref{sec:exp}, we saved a total of four checkpoints at different training steps. We primarily used the final checkpoint, which is the most trained, to compare the three metrics: accuracy, normalized confidence level, and cross-entropy loss. The results for accuracy, normalized confidence level, and cross-entropy loss are presented in Figures \ref{fig:accuracy}, \ref{fig:cl}, and \ref{fig:celoss}.

\textbf{Accuracy}
As opposed to our expectation, the model with the base tokenizer shows higher accuracy than the one with the extended tokenizer, though this difference decreases at larger unit levels: token > character > word, with the hard, word level showing almost the same accuracy for both models. This suggests that if the NTP task targets a larger number of tokens, the accuracy ranking might reverse. The accuracy on the easy level is higher than on the hard level across all unit levels for both the models as predicted. The expected accuracy rankings were token, character, word. However, only the base tokenizer model on the hard level followed this ranking. On the easy level, the base tokenizer model ranked as character, token, word, while the extended tokenizer model ranked as character, word, token on the easy level and character, token, word on the hard level. Notably, the accuracy ranking of the token level for the extended tokenizer model tend to be lower than for the base tokenizer model. This discrepancy likely arises because the extended tokenizer includes a larger vocabulary, where tokens can overlap (e.g., `my' and `mystery'). This overlap can lower accuracy when there are various larger tokens that encompass the target token.

\textbf{Normalized Confidence Level}
As shown in Fig \ref{fig:cl}, the confidence level is consistently lower for the hard level tasks compared to the easy level tasks across all unit levels and both models. This indicates that the models are more certain about their predictions in easier tasks, which aligns with our expectation. The model with the extended tokenizer exhibited higher confidence in the token-level task than the base tokenizer model. However, the opposite trend was observed in the character-level and word-level tasks.

\textbf{Normalized Accurate Confidence Level}
For a more detailed analysis, we further compare the confidence levels in cases where the model predicted correctly and incorrectly, referring to these as the normalized accurate confidence level and normalized inaccurate confidence level, respectively. For the normalized accurate confidence level, the differences between the base tokenizer model and the extended tokenizer model were generally smaller than the overall confidence level difference. In the word-level hard task, ranking even reversed. Additionally, for the character-level task, the base tokenizer model showed higher normalized accurate confidence for the hard task than the easy task. Notably, the extended tokenizer model had a higher normalized accurate confidence level in the hard tasks than in the easy tasks across all three tasks, indicating increased certainty in its predictions for the harder versions.

\textbf{Normalized Inaccurate Confidence Level}
The base tokenizer model showed less normalized inaccurate confidence than the extended tokenizer model in the token-level task. Conversely, for the character-level and word-level tasks, the base tokenizer model exhibited significantly higher normalized inaccurate confidence compared to the extended tokenizer model. This suggests that the base tokenizer model is more likely to produce nonsensical outputs with high certainty, whereas the extended tokenizer model is more cautious in its predictions.

\textbf{Cross Entropy Loss}
Cross entropy loss measures the confusion of the model when generating an output. As expected, both models exhibit higher cross entropy loss in the hard tasks compared to the easy tasks across all unit levels. In the easy tasks, the cross entropy loss of the extended tokenizer model is greater than that of the base tokenizer model. However, in the hard versions of the character-level and word-level tasks, the extended tokenizer model shows lower cross entropy loss. This suggests that the extended tokenizer model performs better in more complex tasks, as character-level and word-level tasks are considered more challenging than token-level tasks.

Summing up the results, while the extended tokenizer model may not necessarily be more accurate than the base tokenizer model in granular tasks, such as predicting a small number of next tokens, it appears to exhibit greater stability. The extended tokenizer model shows less confidence when generating inappropriate tokens and handles difficult tasks with less perplexity compared to the base tokenizer model. This increased stability in generating sensible outputs is likely to positively influence the downstream task performance of language models.

To examine if there are differences in results based on the training steps, we analyzed the changes in accuracy, confidence level, and cross entropy loss at each checkpoint for each task. The results in Appendix \ref{appendix} indicate that after a certain number of training steps, the results tend to converge, suggesting that the results can be generalized at least to TinyLlama based models.

\section{Conclusion} 

In this work, we investigate the influence of language-specifically extended tokenizers on Large Language Models, focusing on a Next Token Prediction Task with two difficulty levels and three unit levels. Using accuracy, confidence level, and cross entropy loss as evaluation metrics, we observe that the extended tokenizer provides more stable generation, exhibiting less confidence in inaccurate token generation and lower cross entropy loss on complex tasks. Our study is the first to explore the internal impact of tokenizers and suggest the possible underlying mechanism of how language-specific tokenizers could enhance downstream task performance.

\section*{Limitation and Future Work}

Since this study experiments with a relatively small LLM, containing only 1.1 billion parameters, the results cannot be directly generalized to larger models. Additionally, as this is a case study on Korean, it is uncertain whether the same results would apply to other languages. We believe that similar intrinsic research on the effect of different tokenizers on LLMs should be further investigated, varying both the sizes of the models and the languages studied.

\bibliographystyle{plainnat}
\bibliography{references} 

\appendix

\section{Appendix}
\label{appendix}

\begin{figure}[h] 
    \centering
    \includegraphics[width=\columnwidth]{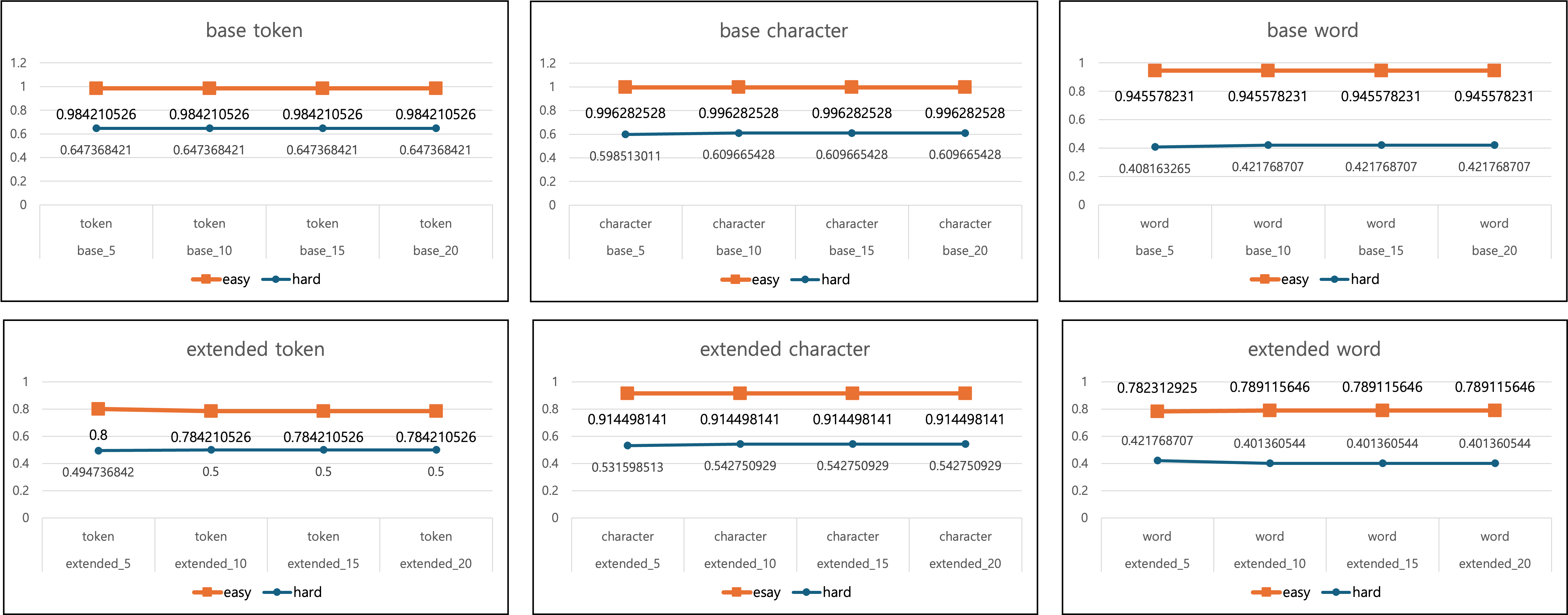} 
    \caption{Accuracy of each checkpoint. The top three rows represent the base tokenizer model, while the bottom three rows represent the extended tokenizer model. The figures on the left show token-level results, the middle figures show character-level results, and the figures on the right show word-level results. Within each figure, the orange points indicate the easy level, and the blue points indicate the hard level. Additionally, within each figure, the leftmost points correspond to the least trained model, and the rightmost points correspond to the most trained model.}
    \label{fig:acc_step}
\end{figure}

\begin{figure}[h] 
    \centering
    \includegraphics[width=\columnwidth]{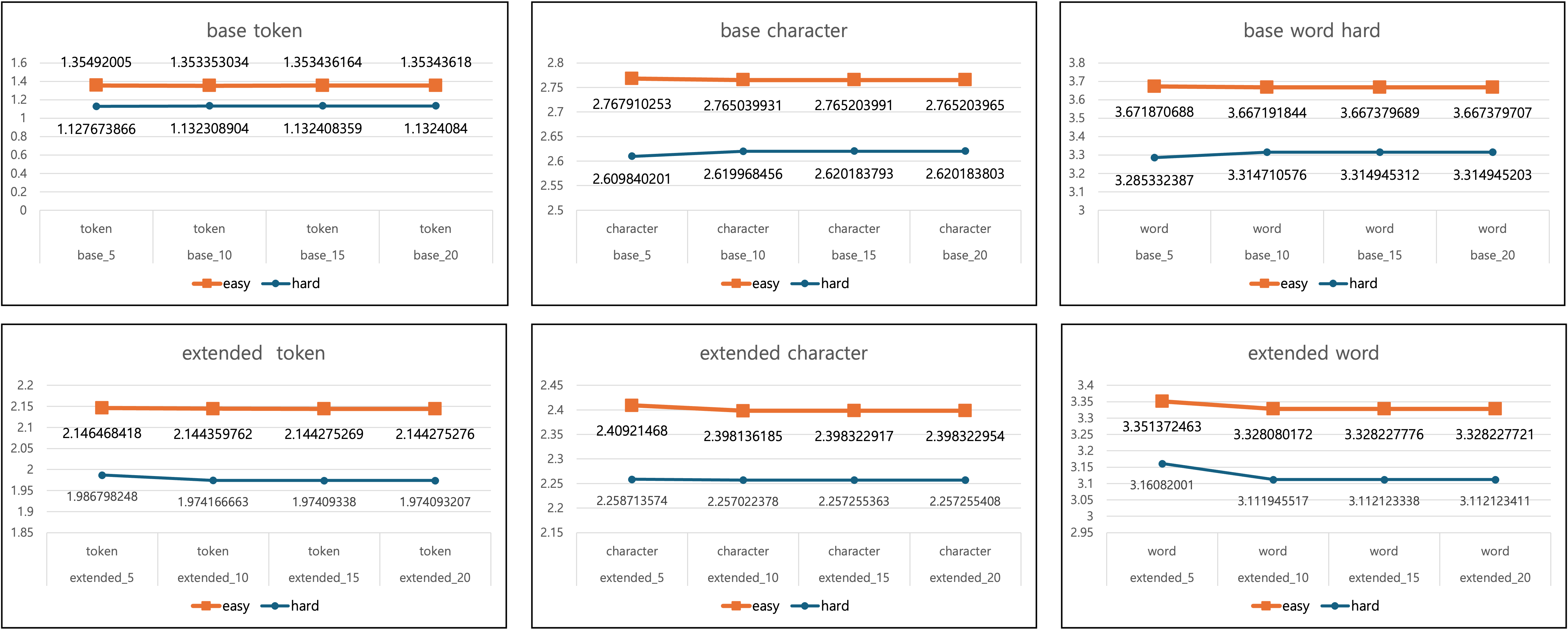} 
    \caption{Normalized confidence level of each checkpoint.}
    \label{fig:cl_step}
\end{figure}

\begin{figure}[h] 
    \centering
    \includegraphics[width=\columnwidth]{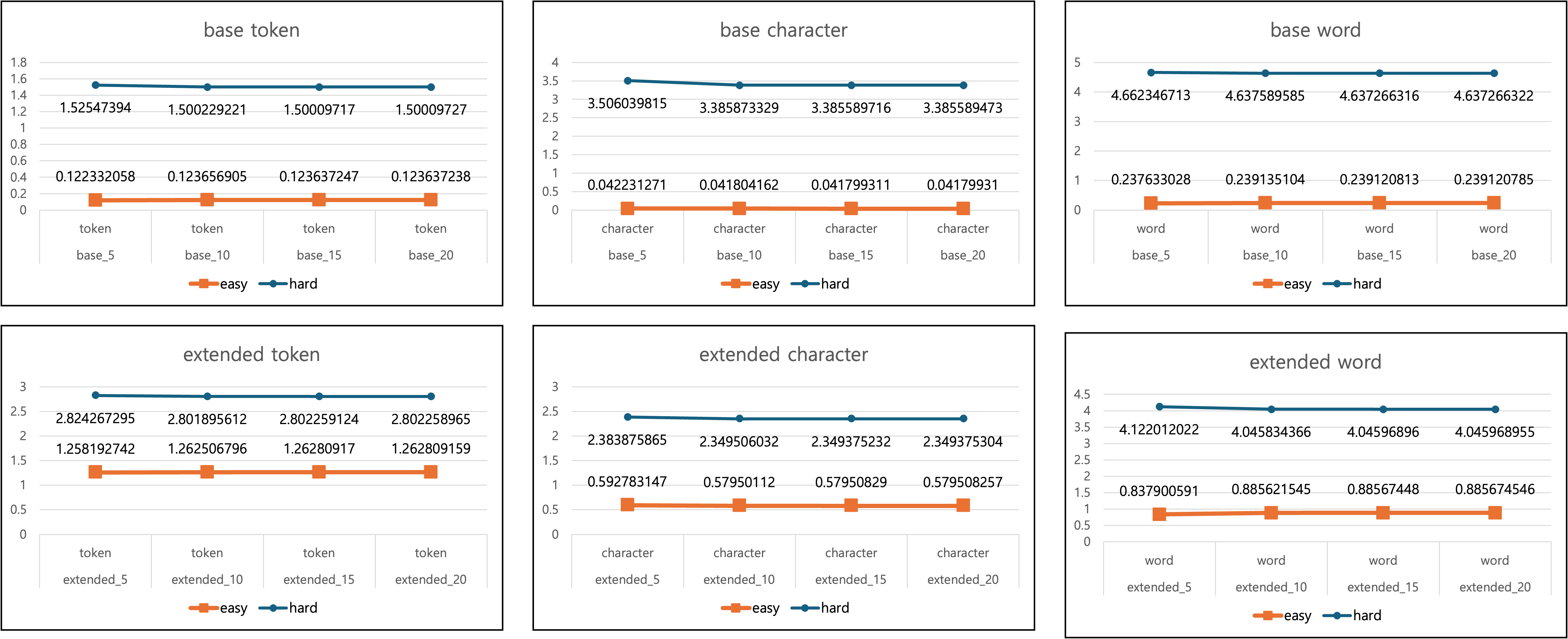} 
    \caption{Cross entropy loss of of each checkpoint.}
    \label{fig:celoss_step}
\end{figure}

\end{document}